\documentclass[letterpaper, 10 pt, conference]{ieeeconf}  % Comment this line out if you need a4paper

\IEEEoverridecommandlockouts                              % This command is only needed if 
\usepackage{graphics} % for pdf, bitmapped graphics files
\usepackage{graphicx}

\usepackage{epsfig} % for postscript graphics files
\usepackage{amsmath} % assumes amsmath package installed
\usepackage{amssymb}  % assumes amsmath package installed
 \usepackage{algorithm}
\usepackage{algpseudocode}
\algnewcommand\AAND{\textbf{ and }}
\algnewcommand\Or{\textbf{ or }}
\usepackage{color}
\usepackage{citesort}
\usepackage{flushend}
\usepackage{url}

\DeclareMathAlphabet{\pazocal}{OMS}{zplm}{m}{n}

\newtheorem{definition}{Definition}

\usepackage{array}
\newcolumntype{C}[1]{>{\centering\arraybackslash}p{#1}}
\newcolumntype{M}[1]{>{\raggedright\arraybackslash}p{#1}}

\usepackage{array} 
\newcolumntype{L}[1]{>{\raggedright\let\newline\\\arraybackslash\hspace{0pt}}m{#1}}	
\newcolumntype{S}[1]{>{\centering\let\newline\\\arraybackslash\hspace{0pt}}m{#1}}
\newcolumntype{R}[1]{>{\raggedleft\let\newline\\\arraybackslash\hspace{0pt}}m{#1}}

\algnewcommand\pushup{\vspace{-1ex}}
\algnewcommand\pushuphalf{\vspace{-0.5ex}}

\newtheorem{problem}{\textbf{Problem}}

\makeatletter
\renewcommand*{\@opargbegintheorem}[3]{\trivlist
  \item[\hskip \labelsep{\itshape #1\ #2}] \textit{(#3)}\ }
\makeatother
\title{\LARGE \bf
Appendix for the Motion Primitives-based Path Planning for Fast and Agile Exploration Method}

\author{Mihir Dharmadhikari, Tung Dang, Kostas Alexis% <-this % stops a space
\thanks{The authors are with the Autonomous Robots Lab, University of Nevada, Reno, 1664 N. Virginia, 89557, Reno, NV, USA
        {\tt\small mdharmadhikari@nevada.unr.edu}}%
}

\begin{document}

\maketitle
\thispagestyle{empty}
\pagestyle{empty}

%%%%%%%%%%%%%%%%%%%%%%%%%%%%%%%%%%%%%%%%%%%%%%%%%%%%%%%%%%%%%%%%%%%%%%%%%%%%%%%%
\begin{abstract}
This manuscript presents enhancements on our motion-primitives exploration path planning method for agile exploration using aerial robots. The method now further integrates a global planning layer to facilitate reliable large-scale exploration. The implemented bifurcation between local and global planning allows for efficient exploration combined with the ability to plan within very large environments, while also ensuring safe and timely return-to-home. A new set of simulation studies and experimental results are presented to demonstrate the new improvements and enhancements. The method is available open-source as a Robot Operating System (ROS) package. 

\end{abstract}

%%%%%%%%%%%%%%%%%%%%%%%%%%%%%%%%%%%%%%%%%%%%%%%%%%%%%%%%%%%%%%%%%%%%%%%%%%%%%%%%
\section{INTRODUCTION}
%%%%%%%%%%%%%%%%%%%%%%%%%%%%%%%%%%%%%%%%%%%%%%%%%%%%%%%%%%%%%%%%%%%%%%%%%%%%%%%%

Research in autonomous robotic exploration and mapping of unknown environments is expanding into an ever increasing set of application domains. Pushing the frontier with respect to the settings and environments within which robots can be utilized as explorers~\cite{grocholsky2006cooperative,KTIO_ICRA_2019,UNDERGROUND_AEROCONF_2019,arora2017randomized,NIR_ICUAS_2017,TUNNEL_AEROCONF_2018}, first responders~\cite{balta2017integrated,tomic2012toward}, and inspectors~\cite{galceran2013survey,SIP_AURO_2015,BABOOMS_ICRA_15,APST_MSC_2015}, aerial vehicles, in particular, are currently employed in a multitude of civilian and military applications. Nevertheless, despite the unprecedented progress in the domain and the multiple exploration strategies proposed~\cite{connolly1985determination,yamauchi1997frontier,popovic2016online,nieuwenhuisen2019search,NBVP_ICRA_16,ROSChapter,VSEP_ICRA_2018,RHEM_ICRA_2017,GBPLANNER_IROS_2019}, the current state-of-the-art, as demonstrated experimentally, is limited to rather low-speed conservative missions as the robots try to guarantee safe navigation and simultaneous optimized selection of subsequent exploration moves given their real-time onboard localization and mapping capabilities. However, low-speed exploration prohibits the exploitation of the full flight envelope and agility of Micro Aerial Vehicles (MAVs) and forbids large-scale exploration given the generally limited battery life available in such platforms. Although historically this was justified due to the limitations of the onboard localization and mapping process, recent progress in robotic perception paves new capabilities for fast and agile exploration if the overall robot autonomy functionality can exploit them. 

Motivated by the discussion above, we recently presented our contribution of motion primitives-based fast exploration path planning using aerial robots. In this appendix we present the extension of the method to allow for global re-positioning towards identified frontiers of the exploration space, as well as to perform auto-homing when the remaining battery necessitates. The description provided is aligned with our open-source contribution available at \url{https://github.com/unr-arl/mbplanner_ros}. 

The rest of this manuscript is organized as follows. Section~\ref{sec:probstat} outlines the full exploration planning problem, followed by an outline of the enhanced planning procedure in Section~\ref{sec:approach}. New simulation studies are presented in Section~\ref{sec:simulation}, while additional experiments focusing on the global planning functionality are detailed in Section~\ref{sec:experimental}. Finally, short conclusions are provided in Section~\ref{sec:concl}.

%%%%%%%%%%%%%%%%%%%%%%%%%%%%%%%%%%%%%%%%%%%%%%%%%%%%%%%%%%%%%%%%%%%%%%%%%%%%%%%%
\section{PROBLEM STATEMENT}\label{sec:probstat}
%%%%%%%%%%%%%%%%%%%%%%%%%%%%%%%%%%%%%%%%%%%%%%%%%%%%%%%%%%%%%%%%%%%%%%%%%%%%%%%%

The problem of agile exploration path planning in subterranean environments, as considered in this work, is an instance of volumetric exploration of unknown space. As such, it aims to build a complete map of an environment for which no prior knowledge exists. Generally, a subterranean environment is a closed, bounded and connected point set with possible exceptions being access points from above ground (e.g., mine portals, subway entrances, doors to facilities with underground floors). It consists of a large-in-scale complex network of branches, junctions, multiple levels and openings (e.g., rooms or caves) and may involve a collection of obstacles within it. As such, these kinds of environments (e.g., underground mines, tunnels, subway infrastructure, cave networks) with their distinct characteristics in length and topology introduce multiple challenges that must be accounted for in the planner. In particular, they are typically very large, and the areas to be navigated through are often confined to tube-like tunnels and branching points. The scale of the environment demands fast exploration and at the same time prohibits planning from taking place in a computationally efficient manner in the full configuration space of the environment bounds. Nevertheless, the identified geometric properties enable a key planning architecture, namely to ``break'' the problem into a ``local'' one, during which the robot searches around its area to find paths that continue its exploration, and a ``global'' one that is triggered when the robot has to backtrack to a previous branching point (or other frontier) to continue its exploration from there. Given this two-layer approach, we define the concepts of ``local completion'' and ``global completion'' for the local and global planners respectively.

Let $\mathbb{M}$ be a $3\textrm{D}$ occupancy map of the environment which is incrementally built from measurements of an onboard depth sensor $\mathbb{S}$, as well as robot poses derived from a localization system $\mathbb{O}$ responsible to fuse exteroceptive and proprioceptive data to simultaneously estimate the robot path and the map of its environment. The map consists of voxels $m$ of three categories, namely $m \in \mathbb{M}_{free}$, $m \in \mathbb{M}_{occupied}$, or $m\in \mathbb{M}_{unknown}$ representing free ($V_{free}$), occupied ($V_{occ}$), and unknown ($V_{unm}$) space respectively. Furthermore, let $d_{\max}$ be the effective range, and $[F_H, F_V]$ be the field-of-view in horizontal and vertical directions respectively of the depth sensor $\mathbb{S}$. In addition, let the robot's configuration at time $t$ be defined as the combination of $3\textrm{D}$ position, linear velocity and heading $\xi_t =[x_t,y_t,z_t,v_x,v_y,v_z,\psi_t]$. Importantly, since for most range sensors' perception stops at surfaces, sometimes hollow spaces or narrow pockets cannot be fully explored thus leading to a residual map $\mathbb{M}_{\ast, res} \subset \mathbb{M}_{unknown}$ with volume $V_{\ast,res}$ which is infeasible to explore given the robot's constraints. As a result, given a volume $V_{\ast}$, the potential volume to be explored is $V_{\ast,explored} = V_{\ast} \setminus V_{\ast,res}$.

\begin{definition}[Local Completion]
\label{def:localcompletion}
Given a map $\mathbb{M}$, within a local sub-space $\mathbb{M}_L$ of dimensions $D_L$ centered around the current robot configuration, the planner reports \textit{``local completion''} if $V_{D_L,explored} = V_{D_L} \setminus V_{D_L,res}$. % $V_{explored} = V_{D_L} \setminus V_{D_L,res}$
\end{definition}

\begin{definition}[Global Completion]
\label{def:globalcompletion}
Given the full occupancy map $\mathbb{M}$ of the environment with dimensions $D_G$ and volume $V_{D_G}$, the planner considers \textit{``global completion''} if $V_{D_G,explored} = V_{D_G} \setminus V_{D_G, res}$.
\end{definition}

In practice, it is not possible and unrealistic to identify $V_{res}$, but completion can be approximated by the lack of a collision-free path inside a planning volume which leads to a space with potentially unknown volume larger than a threshold $V_\delta$. The local and global planner problems are formulated as follows. 

\begin{problem}[Local Exploration Planner]\label{prob:localplanner}
Given an occupancy map $\mathbb{M}$ and a local subset of it $\mathbb{M}_L$ centered around the current robot configuration $\xi_{0}$, find a collision-free and traversability-aware (when applicable) path $\sigma_L=\{\xi_i\}$ to guide the robot towards unmapped areas and maximize an exploration gain defined as the volume which is expected to be mapped when the robot traverses along the path $\sigma_L$ with a sensor $\mathbb{S}$. A path is admissible if it is collision-free in the sense of not colliding with $3\textrm{D}$ obstacles in the map and respecting the robot dynamic model. When ``local completion'' is reported by this planner, the global planner is to be engaged. 
\end{problem}

\begin{problem}[Global Exploration Planner]\label{prob:globalplanner}
Given the explored and unknown subsets of an occupancy map $\mathbb{M}$ of the environment and the current robot configuration $\xi_{0}$, find a dynamics-aware and collision-free path $\sigma_G$ leading the robot towards the frontiers of the unmapped areas. Feasible paths of this planning problem must take into account the remaining endurance of the robot. When the environment is explored completely (``global completion'') or the battery limits are approaching, find a collision-free path $\sigma_{H}$ to return the robot to its home location $\xi_{home}$.
\end{problem}

%%%%%%%%%%%%%%%%%%%%%%%%%%%%%%%%%%%%%%%%%%%%%%%%%%%%%%%%%%%%%%%%%%%%%%%%%%%%%%%%
\section{METHOD ENHANCEMENTS}\label{sec:approach}
%%%%%%%%%%%%%%%%%%%%%%%%%%%%%%%%%%%%%%%%%%%%%%%%%%%%%%%%%%%%%%%%%%%%%%%%%%%%%%%%

\begin{figure}[h!]
\centering
    \includegraphics[width=0.99\columnwidth]{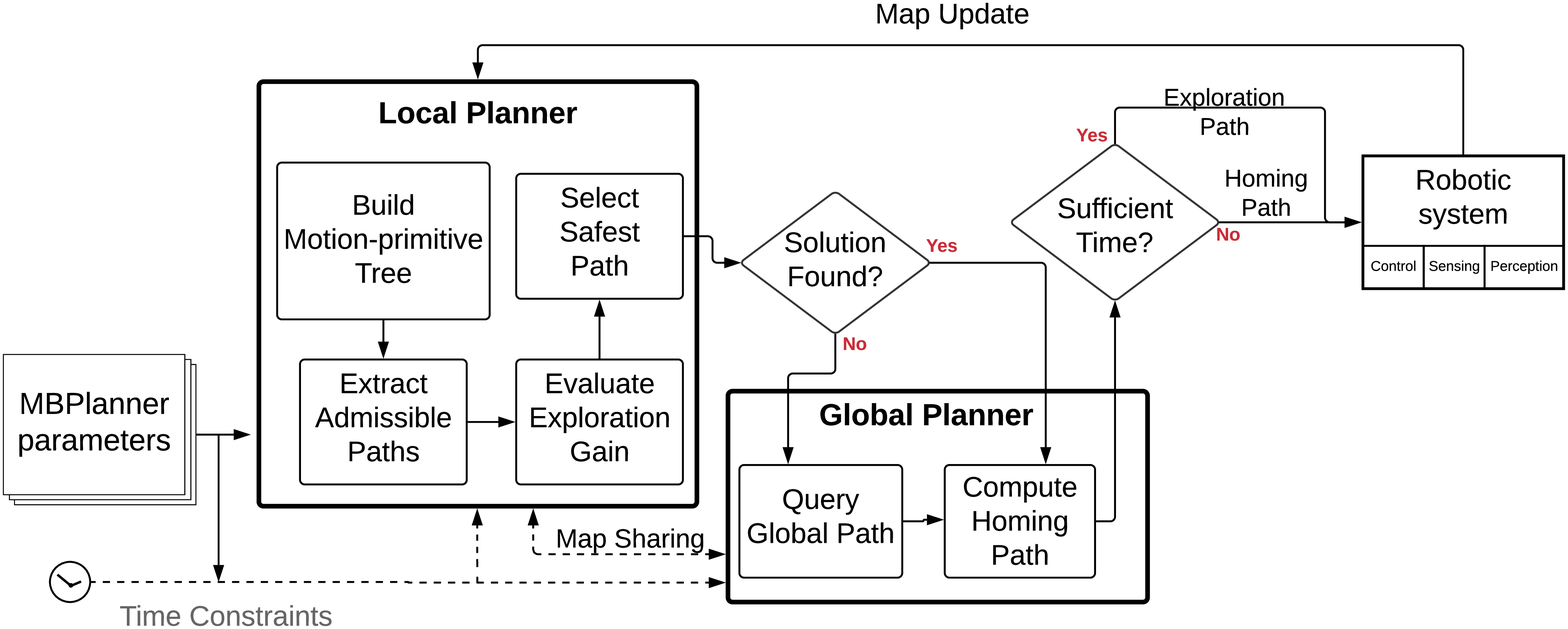}
\caption{Block diagram view of the proposed planner architecture, called MBPlanner, involving the local layer of motion primitives-based exploration and its global layer responsible for auto-homing and repositioning towards frontiers of the exploration space when a successful local exploration path is not possible to be identified.}\label{fig:PlannerArchitecture}
%\vspace{-2ex}
\end{figure}

Subterranean settings typically involve long and confined corridors connected by multi-way intersections. Such environments therefore correspond to major challenges for robotic exploration due to the scale (e.g., km in length) and geometric complexity. In response to these facts, this work proposes a bifurcated local-global exploration path planning method, called ``Motion primitivies-based exploration path planner'' (MBPlanner), that at its core exploits motion primitives-based planning to identify fast and agile paths, while also detecting frontiers of the explored subset of the environment towards which the robot can backtrack when needed to continue its exploratory mission. The method is an improved version of our previous publication on MBPlanner~\cite{MBPlanner_ICRA_2019} and is now released as an open-source contribution. In further detail, the local motion primitives-based exploration planner, responsible for finding paths to maximize the amount of volume mapped, searches within a local space of fixed dimension to enable fast computation and the derivation of agile and efficiently exploring trajectories that exploit the aggressive dynamics of MAVs. The exploration planner plans paths in $3\textrm{D}$ enabling exploration of environments with varying altitude. Due to the geometric particularities of underground settings in terms of scale, complexity and topology, the local planner might reach a dead-end or other scenario that prohibits the derivation of an effective exploration path. The method is thus extended to incorporate a global planner used to derive a) a safe and timely return-to-home path, and b) paths towards previously identified frontiers of the exploration space. An example use of the global planner is that after a robot has explored the main drift of the underground mine, it reaches a dead-end (mine heading) and thus has to return to a previously found branching point in order to continue its mission. The overall diagram of the proposed solution is shown in Figure~\ref{fig:PlannerArchitecture}. The local planning mode is detailed in~\cite{MBPlanner_ICRA_2019}, whereas the global planning functionality is analogous to the one presented in our previous contribution on subterranean exploration~\cite{GBPLANNER_JFR_2020}. 

% that at its core exploits motion primitives-based planning to identify fast and agile paths, while also detecting frontiers of the explored subset of the environment towards which the robot can backtrack when needed to continue its exploratory mission.
%\vspace{-2ex}
%
%%%%%%%%%%%%%%%%%%%%%%%%%%%%%%%%%%%%%%%%%%%%%%%%%%%%%%%%%%%%%%%%%
%%%%%%%%%%%%%%%%%%%%%%%%%%%%%%%%%%%%%%%%%%%%%%%%%%%%%%%%%%%%%%%%%
%

%%%%%%%%%%%%%%%%%%%%%%%%%%%%%%%%%%%%%%%%%%%%%%%%%%%%%%%%%%%%%%%%%%%%%%%%%%%%%%%%
\section{SIMULATION STUDIES}\label{sec:simulation}
%%%%%%%%%%%%%%%%%%%%%%%%%%%%%%%%%%%%%%%%%%%%%%%%%%%%%%%%%%%%%%%%%%%%%%%%%%%%%%%%

\begin{figure}[h!]
\centering
    \includegraphics[width=0.99\columnwidth]{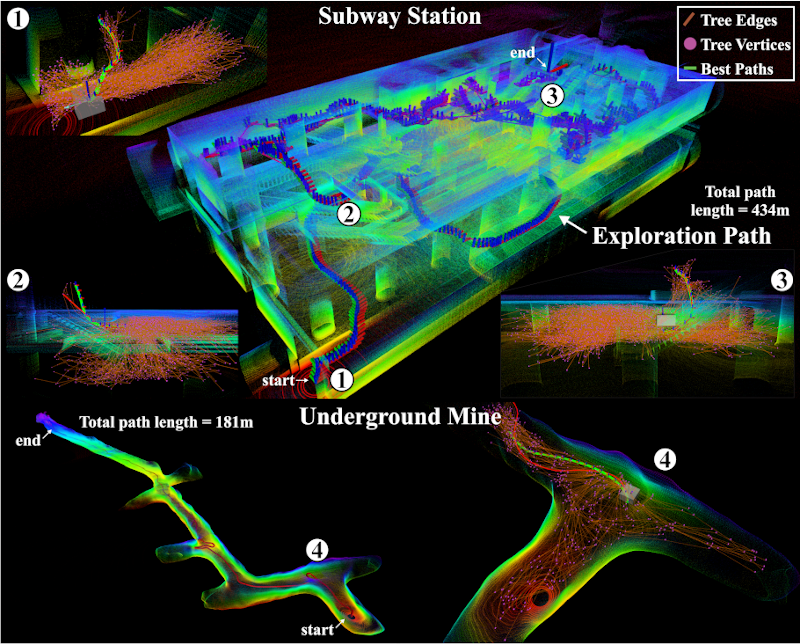}
\caption{Simulation-based evaluation of the local stage of MBPlanner both in a subway station and in an underground mine. Selected paths: 1) take-off and initial maneuver, 2) ascending to the next level through the opening of a staircase, 3) also ascending to the next level, 4) handling an intersection inside a mine. This study focuses on the more complex $3\textrm{D}$ exploration capabilities of the method.} \label{fig:simulationresult}
% \vspace{-2ex}
\end{figure}

A set of simulation studies were conducted in order to evaluate and fine-tune the enhanced motion primitives-based exploration planner and its local and global planning layers. First of all, the local exploratory planning stage was evaluated through two simulation studies inside a) a subway station, and b) an underground mine. The simulation studies were conducted using the RotorS Simulator~\cite{furrer2016rotors}, while the local planning window of the planner (the volume within which the motion primitives-based tree is built) is set to $D_L$ of \textit{length}$\times$\textit{width}$\times$\textit{height} = $40\times 40\times 8\textrm{m}$ and the robot bounding box $D_R$ is considered equal to \textit{length}$\times$\textit{width}$\times$\textit{height} = $0.6\times 0.6\times 0.6\textrm{m}$. Both simulation studies were conducted assuming a quadrotor micro aerial vehicle model integrating a LiDAR sensor with $[F_H,F_V]=[360,30]^\circ$ and $d_{\max}=50\textrm{m}$ and the exploration speed was set to $2\textrm{m/s}$. This test, depicted in Figure~\ref{fig:simulationresult}, indicates the ability of the proposed method to explore both large-scale environments such as underground mines and buildings such as the subway station. Note that the subway station contained multiple levels requiring going up floors through the openings in stairways demonstrating the $3\textrm{D}$ exploration capability of the method.

%
%%%%%%%%%%%%%%%%%%%%%%%%%%%%%%%%%%%%%%%%%%%%%%%%%%%%%%%%%%%%%%%%%
%%%%%%%%%%%%%%%%%%%%%%%%%%%%%%%%%%%%%%%%%%%%%%%%%%%%%%%%%%%%%%%%%
%

A third simulation study was further conducted with the goal to evaluate and verify the whole global and local planning functionalities of the proposed planner. In this simulation study the environment used was an underground room-and-pillar mine. This room-and-pillar mine environment is multiple km-long and presents an array of multi-way intersections that challenge the planner’s behavior. The performance of the proposed planner is compared against the receding horizon Next-Best-View Planner (NBVP)~\cite{NBVP_ICRA_16} and the Frontiers exploration algorithm (FrontierPlanner)~\cite{yamauchi1997frontier} implemented for 3D environments and combined with an optimal sampling-based motion planner for collision-free navigation to the frontiers~\cite{RRTS1a}.

\begin{figure}
\centering
    \includegraphics[width=0.99\columnwidth]{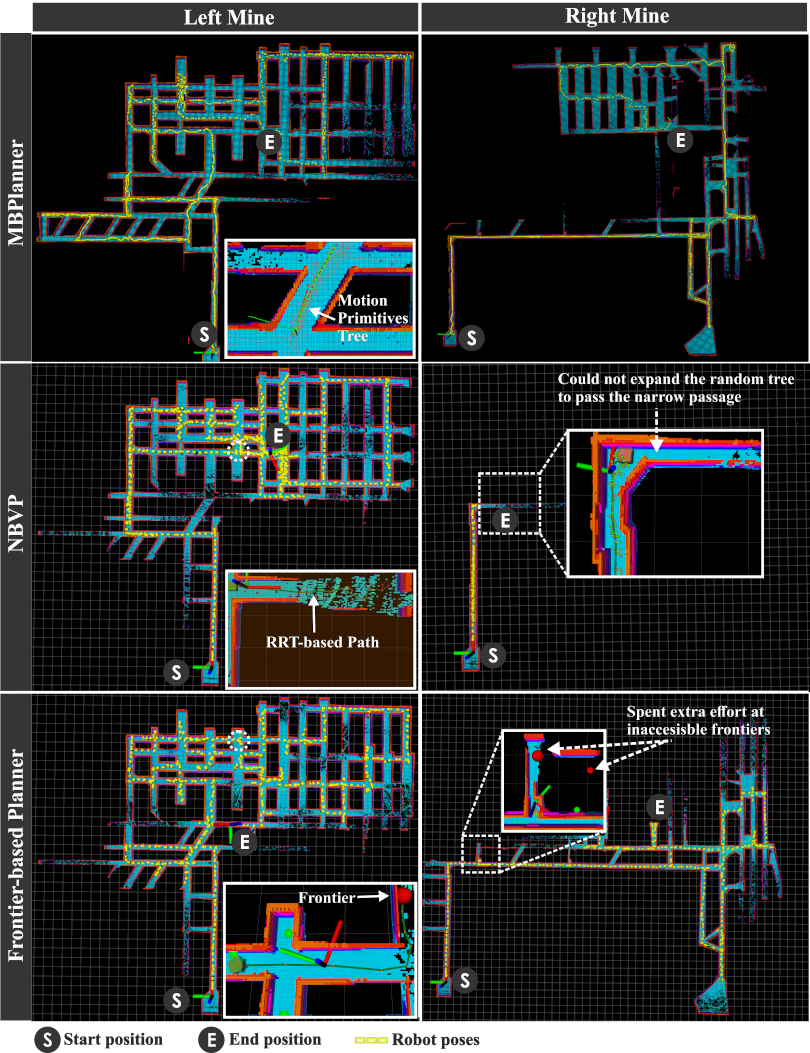}
\caption{Simulation results for autonomous exploration inside a room-and-pillar mine. The robot is tasked to explore either the left or the right section of the environment. The left column shows exploration maps and indicative planning examples from each planner in the left mine. In general, they all demonstrate promising performance in this environment, mainly due to its wide corridors and presence of multiple nearby intersections for exploration at any location. As shown, the mission with MBPlanner was almost complete, while NBVP tends to provide solutions biased towards spacious corridors since its samples are spanned over the whole map. The FrontierPlanner achieves adequate performance but with the lowest exploration rate because it does not account for the range sensor model as MBPlanner and NBVP do. In the right column of the figure, indicative exploration results from each planner are depicted. The MBPlanner outperforms the other two methods. The FrontierPlanner struggles with inaccessible frontiers detected inside narrow areas, while NBVP gets trapped in the first corridor since its random tree, expanding over the whole map, is not dense enough to pass through the first narrow passage.}\label{fig:roomandpillarsim}
% \vspace{-2ex}
\end{figure}

\begin{figure}[h!]
\centering
    \includegraphics[width=0.99\columnwidth]{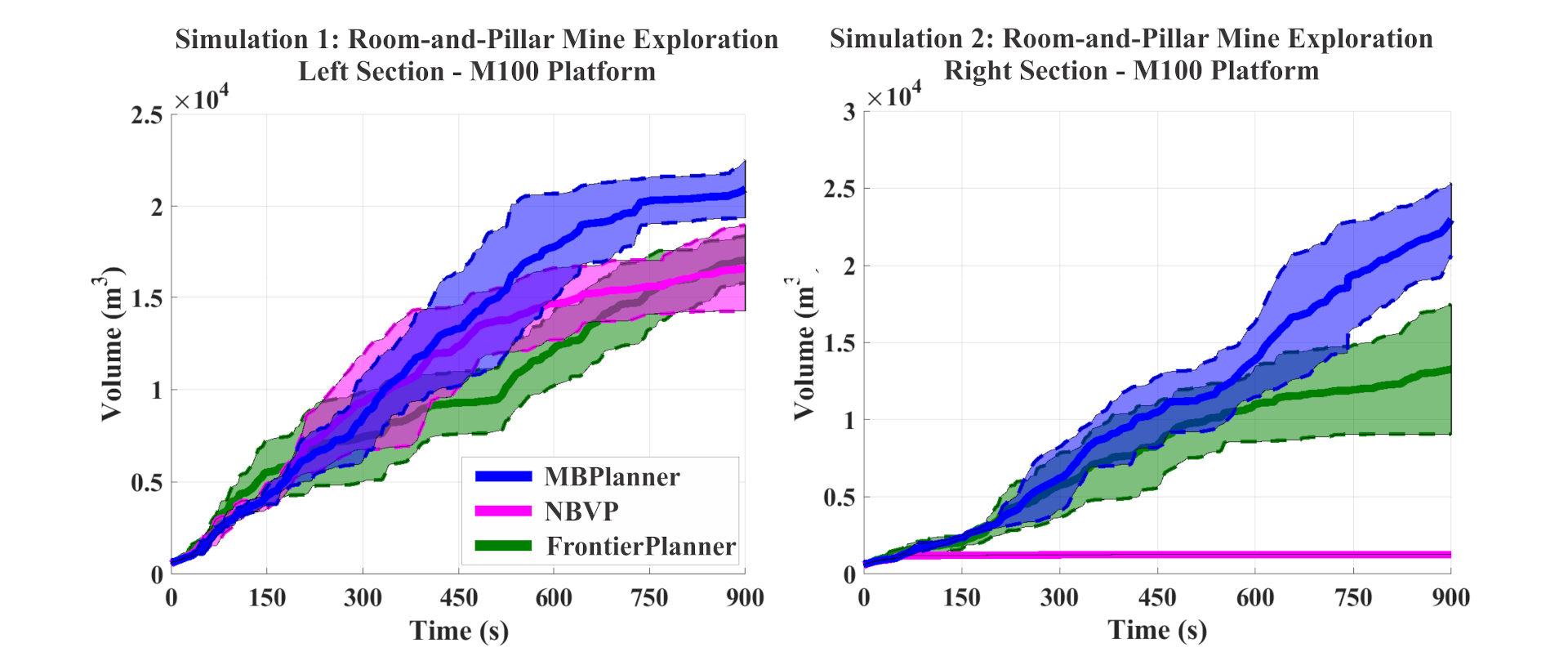}
\caption{Comparison of exploration progress from the three considered planners (MBPlanner, NBVP, and FrontierPlanner) in a simulated room-and-pillar mine. The left and the right sub-figures show the changes in the exploration volume during five 15min independent runs inside the left and the right section of the mine respectively. The solid lines present the average over the whole five runs associated with shaded areas which are the lower and upper bound from those runs. Overall, MBPlanner outperforms the other two methods and works reliably in both cases. NBVP is incapable of providing planning solutions for exploring narrow spaces due to its fixed–global sampling space setting, thus leading to an insufficient number of samples to find feasible paths through narrow corridors within reasonable computational bounds. Furthermore, the FrontierPlanner achieves slower exploration rate and is sensitive to inaccessible frontiers that arise in tight spaces.}\label{fig:roomandpillarsim_matlab}
% \vspace{-2ex}
\end{figure}

The room-and-pillar simulated environment consists of two sections, left and right, with the first being relatively more spacious with respect to the width of its corridors and the second rather more constrained. Hence, for a fair comparison between these three planners, two exploration scenarios are employed corresponding to the left and the right sections of the mine. A quadrotor model, similar to a real robotic system is developed in ROS-Gazebo utilizing the RotorS simulator~\cite{furrer2016rotors}. In both sections of the mine, each planner is run 5 times for 15 minutes each (a typical flight time for our flying platform). The average velocity of the robot is set to $2\textrm{m/s}$. Relevant simulation results presenting the performance of a) MBPlanner, b) NBVP, c) FrontierPlanner in the left and right subsets of this environment are shown in Figure~\ref{fig:roomandpillarsim}. Furthermore, the statistical comparison data with respect to the exploration rates of each method are depicted in Figure~\ref{fig:roomandpillarsim_matlab}. As it can be seen, MBPlanner outperforms the other two approaches. More specifically, in the left mine, all three planners present good exploration behavior, which could be attributed to the fact that the environment comprises of rather spacious tunnels and multiple intersections hence easier to find efficient exploration paths. Nevertheless, the MBPlanner achieves the highest exploration rate on average thanks to its fast and smooth planning trajectories. Regarding the right mine, the MBPlanner provides a superior performance compared to the other two methods. NBVP gets trapped at the first narrow passage since its random tree is too sparse as it spreads over the whole explored map. The FrontierPlanner fails to provide comparable results because it spends unnecessary effort trying to reach inaccessible frontiers detected inside tight spaces.

%\vspace{-2ex}
%
%%%%%%%%%%%%%%%%%%%%%%%%%%%%%%%%%%%%%%%%%%%%%%%%%%%%%%%%%%%%%%%%%
%%%%%%%%%%%%%%%%%%%%%%%%%%%%%%%%%%%%%%%%%%%%%%%%%%%%%%%%%%%%%%%%%
%

%
%%%%%%%%%%%%%%%%%%%%%%%%%%%%%%%%%%%%%%%%%%%%%%%%%%%%%%%%%%%%%%%%%
%%%%%%%%%%%%%%%%%%%%%%%%%%%%%%%%%%%%%%%%%%%%%%%%%%%%%%%%%%%%%%%%%
%

%%%%%%%%%%%%%%%%%%%%%%%%%%%%%%%%%%%%%%%%%%%%%%%%%%%%%%%%%%%%%%%%%%%%%%%%%%%%%%%%
\section{EXPERIMENTAL EVALUATION}\label{sec:experimental}
%%%%%%%%%%%%%%%%%%%%%%%%%%%%%%%%%%%%%%%%%%%%%%%%%%%%%%%%%%%%%%%%%%%%%%%%%%%%%%%%

Alongside the main experimental studies presented in~\cite{MBPlanner_ICRA_2019} that involved a collision-tolerant flying robot, in this appendix we present additional experiments that among others focus on the global planning behavior. In these experiments we utilized a different aerial robotic platform to evaluate the combination of the local and global planning stages of the proposed exploration architecture. The utilized autonomous flying robot is developed around a DJI Matrice M100 and integrates a multi-modal sensor fusion solution combined with loosely-coupled LiDAR Odometry And Mapping, as well as visual-inertial localization. The system relies on Model Predictive Control (MPC) for its automated operation. The proposed planner subscribes to the data provided by the localization and mapping solution and provides references to the onboard MPC. Details for the overall system solution can be found in~\cite{UNDERGROUND_AEROCONF_2019,Thermal_ICUAS_2018,mpc_rosbookchapter,KTIO_ICRA_2019}. The depth sensor integrated on the platform is a Velodyne PuckLITE which provides a horizontal and vertical field of view of $F_H=360^{\circ},~F_V=30^{\circ}$. It also has a maximum range of $100\textrm{m}$, while a map update takes place only for the first $50\textrm{m}$ of ranging. Based on the robot size and safety considerations, the bounding box for this test was set to \textit{length}$\times$\textit{width}$\times$\textit{height} = $1.4\times 1.4\times 0.5\textrm{m}$. The system is first deployed and tested inside corridors of the Applied Research Facility (ARF) building at the University of Nevada, Reno. This environment emulates an underground setting with narrow passages leading to dead-end situations and also a branching point which overall requires the global planning behaviour. Another field experiment is conducted inside one of the Truckee River abandoned railroad tunnels which were historically used to connect California with Nevada and towards the East Coast of the U.S. by train. This specific tunnel system is relatively straight but involves significant height and interesting rock-wall geometry which, as it turned out, provided sufficient edges and broadly, geometric features to constraint the localization and mapping process.

%
%%%%%%%%%%%%%%%%%%%%%%%%%%%%%%%%%%%%%%%%%%%%%%%%%%%%%%%%%%%%%%%%%
\begin{figure}
\centering
    \includegraphics[width=0.99\columnwidth]{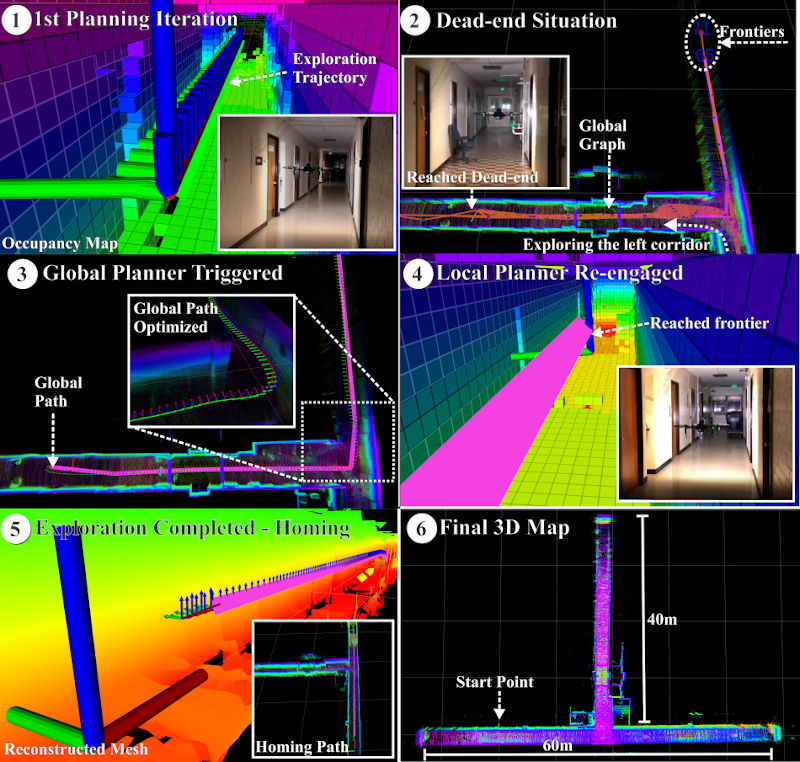}
\caption{Autonomous exploration inside corridors of the Applied Research Facility at the University of Nevada, Reno. The environment consists of two main narrow corridors and a three-way intersection. This environment is suitable to test the local planning in narrow settings, as well as the global re-positioning and auto-homing behavior. The robot was initialized inside one corridor, progressed through the first straight corridor then turned to the left branch. It continued exploring the left corridor until reaching its end point (sub-figures 1-2). The global planner was then triggered to provide a repositioning path back to the first corridor, followed by a refinement step to smooth the path (sub-figure 3-4). Finally, the homing operation was engaged after the robot finished the exploration (sub-figure 5). The travelled distance is about $120\textrm{m}$ in total during the $4$-minute flight. A video of this experimental can be found at \protect\url{https://youtu.be/_z2Sc8ANQa8?t=125}. }\label{fig:arfresult}
%\vspace{-2ex}
\end{figure}
%%%%%%%%%%%%%%%%%%%%%%%%%%%%%%%%%%%%%%%%%%%%%%%%%%%%%%%%%%%%%%%%%
%

%
%%%%%%%%%%%%%%%%%%%%%%%%%%%%%%%%%%%%%%%%%%%%%%%%%%%%%%%%%%%%%%%%%
\begin{figure}[h]
\centering
    \includegraphics[width=0.99\columnwidth]{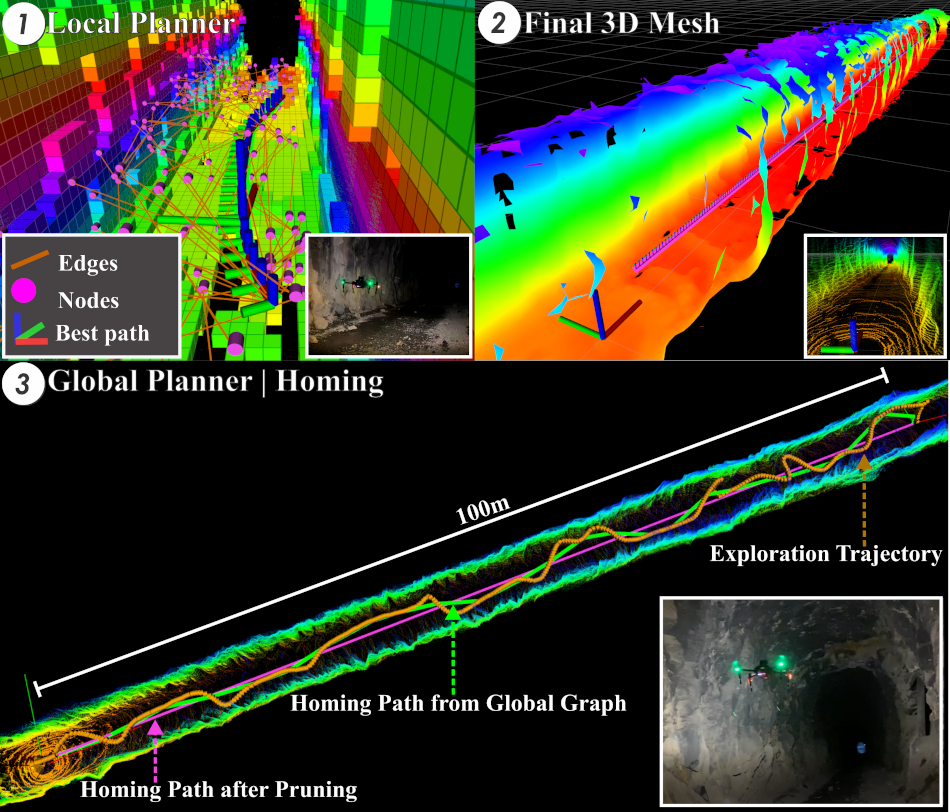}
\caption{Autonomous exploration mission inside the Truckee Railroad Tunnel, California. This abandoned train tunnel is mostly straight and has dimensions of $500\times 5 \times 7\textrm{m}$. The robot, based on a Matrice M100 platform, was initialized inside the tunnel and commanded to explore the tunnel for $5$ minutes with a nominal speed of $0.75\textrm{m/s}$. The local planner provided smooth paths for rapid exploration and the global planner triggered the homing procedure at the end. In total, the robot was able to explore around $100\textrm{m}$ and came back safely to the initial starting location. A video of this experimental can be found at \protect\url{https://youtu.be/_z2Sc8ANQa8?t=236}. }\label{fig:truckeeresult}
%\vspace{-2ex}
\end{figure}
%%%%%%%%%%%%%%%%%%%%%%%%%%%%%%%%%%%%%%%%%%%%%%%%%%%%%%%%%%%%%%%%%
%

In the first test, the robot took-off inside one corridor of the ARF building ($2\textrm{m}$ in width and $2.5\textrm{m}$ in height) and progressively explored the left branch until reaching the first dead-end. The global planner was then queried in order to re-position the robot back to the main corridor to continue the exploration. The global path from the graph is further refined to provide a smoother trajectory respecting the vehicle dynamics. Finally the homing procedure was triggered once the robot completely explored all corridors in the defined bounds. The traveled distance for the whole mission is approximately $120\textrm{m}$ in total with the average speed being set to $0.5\textrm{m/s}$. As shown in Figure~\ref{fig:arfresult}, the robot was able to complete the exploration mission utilizing both the local and global planner, and safely return to the initial starting location. 

In the second experiment, inside the Truckee train tunnel, the robot is tasked to explore the tunnel with a speed setpoint of $0.75\textrm{m/s}$ and must return to the home location automatically given a limited $5$ minutes flight time. The robot was deployed inside the tunnel, autonomously exploring the main tunnel using the local planner until its remaining time approaches the limit. At this point the global planner was automatically invoked to provide the shortest homing path. It is noted that, different to previous test environments, this tunnel is quite tall (approximately $7.5\textrm{m}$ high) which requires the local planner to vary the robot's height to map the whole environment. The Figure~\ref{fig:truckeeresult} depicts the exploration process. The robot traveled a total of about $200\textrm{m}$ in $5$ minutes.

The complete set of these field experiments, alongside those presented in~\cite{MBPlanner_ICRA_2019}, demonstrates the real-life potential of the proposed planner to explore complex subterranean environments using aerial robots. Especially when combined with a collision-tolerant design (see~\cite{MBPlanner_ICRA_2019}) it enables the fast and agile exploration which in turn allows to map large-scale and narrow settings. The latter is particularly important for subterranean environments which may present several challenges to the perception system. 

%%%%%%%%%%%%%%%%%%%%%%%%%%%%%%%%%%%%%%%%%%%%%%%%%%%%%%%%%%%%%%%%%%%%%%%%%%%%%%%%
\section{CONCLUSIONS}\label{sec:concl}
%%%%%%%%%%%%%%%%%%%%%%%%%%%%%%%%%%%%%%%%%%%%%%%%%%%%%%%%%%%%%%%%%%%%%%%%%%%%%%%%

In this short appendix we outlined the architecture of the enhanced local-global search policy of the ``Motion primitivies-based exploration path planner'' (MBPlanner) method which is now available open-source. A set of simulation and experimental studies present the performance of the strategy, its true $3\textrm{D}$ exploration capabilities and the utilization of the global planning mode to re-position the robot towards previously identified frontiers of the exploration space. Last but not least, autonomous homing operation is also provided and ensures that the robot self-commands a homing trajectory when its remaining battery life necessitates. The code related to this work is released as an open-source ROS package (\url{https://github.com/unr-arl/mbplanner_ros}).

%%%%%%%%%%%%%%%%%%%%%%%%%%%%%%%%%%%%%%%%%%%%%%%%%%%%%%%%%%%%%%%%%%%%%%%%%%%%%%%%
\bibliographystyle{IEEEtran}
\bibliography{MBPlanner_Appendix}
%%%%%%%%%%%%%%%%%%%%%%%%%%%%%%%%%%%%%%%%%%%%%%%%%%%%%%%%%%%%%%%%%%%%%%%%%%%%%%%%

\end{document}